\documentclass[sigconf]{aamas}

\usepackage{commands}

\usepackage{balance} 
\usepackage[title]{appendix}
\usepackage{listings}
\usepackage{multirow}
\usepackage{graphicx}
\usepackage{booktabs} 
\usepackage{algorithm}
\usepackage{algpseudocode}
\usepackage{xurl}
\setcopyright{ifaamas}
\acmConference[AAMAS '26]{Proc.\@ of the 25th International Conference
on Autonomous Agents and Multiagent Systems (AAMAS 2026)}{May 25 -- 29, 2026}
{Paphos, Cyprus}{C.~Amato, L.~Dennis, V.~Mascardi, J.~Thangarajah (eds.)}
\copyrightyear{2026}
\acmYear{2026}
\acmDOI{}
\acmPrice{}
\acmISBN{}

\acmSubmissionID{17}

\title[AAMAS-2026 Formatting Instructions]{Distilling Game Code World Model Generation into Lightweight Large Language Models}

\author{Tyrone Serapio}
\affiliation{
  \institution{Brown University}
  \city{Providence}
  \country{United States}}
\email{tyrone_kirk_serapio@brown.edu}

\author{Arjun Prakash}
\affiliation{
  \institution{Brown University}
  \city{Providence}
  \country{United States}}
\email{arjun_prakash@brown.edu}

\author{Haoyang Xu}
\affiliation{
  \institution{Brown University}
  \city{Providence}
  \country{United States}}
\email{haoyang_xu@brown.edu}

\author{Kevin Wang}
\affiliation{
  \institution{Brown University}
  \city{Providence}
  \country{United States}}
\email{kevin_a_wang@brown.edu}

\author{Amy Greenwald}
\affiliation{
  \institution{Brown University}
  \city{Providence}
  \country{United States}}
\email{amy@brown.edu}

\begin{abstract}
Large Language Models (LLMs) have shown great ability in generating executable code from natural language, opening the possibility of automatically constructing environments for AI agents. Recent work on Code World Models (CWMs) demonstrates that LLMs can translate game rules into Python implementations compatible with solvers like Monte Carlo Tree Search. We study this problem in game settings, where generated environments must implement rules, legal actions, state transitions, observations, and rewards. We refer to these game-specific executable models as Game Code World Models (GameCWMs). However, current approaches to generating code world models rely on frontier models and inference-time refinement loops, limiting accessibility and scalability. This work investigates whether GameCWM generation capabilities can be distilled into smaller models through post-training. We introduce: (1) a curated dataset of 30 games spanning perfect and imperfect information games, (2) a verification framework that evaluates generated code against structural and semantic game properties, and (3) a post-training pipeline combining Supervised Fine-Tuning (SFT) with \styrone{Group Relative Policy Optimization (GRPO)}{Reinforcement Learning with Verifiable Rewards (RLVR)}. We experiment with Qwen2.5-3B-Instruct and find that SFT can increase syntactic correctness, while \styrone{GRPO}{RLVR} can improve execution-level adherence to game rules, thereby improving Qwen's ability to generate valid GameCWMs in both perfect and imperfect information games. Overall, our pipeline makes Qwen2.5-3B-Instruct more capable of generating valid GameCWMs, thereby offering a scalable path toward automatic environment generation from natural language.
\end{abstract}

\begin{document}



         
\newcommand{\BibTeX}{\rm B\kern-.05em{\sc i\kern-.025em b}\kern-.08em\TeX}

\newcommand{\sstate}{s}
\newcommand{\player}{p}
\newcommand{\observation}{O_p}
\newcommand{\hhistory}{h}
\newcommand{\action}{a}
\newcommand{\Score}{S}
\newcommand{\gameprompt}{P_G}
\newcommand{\gamecode}{C_G}
\newcommand{\weight}{w}
\newcommand{\game}{G}



\pagestyle{fancy}
\fancyhead{}


\maketitle 

\section{Introduction}

A world model is an AI's internal representation of the world with which it is designed to interact.
It is a description of how the AI's environment works, which enables the AI to reason before it acts by simulating the effects of its actions internally.
While world models are immensely valuable once deployed, building them can be an onerous task.
As a result, recent research has been geared towards automating the building process.

One straightforward method is to assume that LLMs are sufficiently capable to serve as world models themselves.
With this approach, you simply ask the LLM to weigh the consequences of its future actions before taking any. 
This approach has not been particularly successful to date, especially in long-horizon tasks, as LLMs tend to
generate unworkable plans \cite{rana2025modelfirstreasoningllmagents}.
Moreover, their performance degrades in out-of-distribution environments \cite{chen2025internalizingworldmodelsselfplay}.

An alternative is to give an LLM a natural-language description of an environment, and ask it to generate code that faithfully models the physical world as described \cite{deng2025naturallanguageextensiveformgame, lehrach2025codeworldmodelsgeneral, wang2026generalizableframeworkbuildingexecutable}.
As long as we can verify the faithfulness of the world model to the given environment, this latter approach should prove more reliable and interpretable
\cite{dainese2024generatingcodeworldmodels}. 
To this end, \citet{dainese2024generatingcodeworldmodels} introduced the concept of Code World Models (CWMs) to construct executable Python environments for single-agent reinforcement learning environments.

Today's world models are primarily designed for a single agent, and even when additional agents are present in the environment, all but one are usually assumed to be passive: e.g., in Mujoco \cite{todorov2012mujoco}.%
\footnote{One notable exception is multi-robot model predictive controllers \cite{tajbakhsh2024conflictbasedmodelpredictivecontrol}.} 
Building on \citeauthor{dainese2024generatingcodeworldmodels}'s work, \citet{lehrach2025codeworldmodelsgeneral} extended CWMs to multi-agent game-theoretic environments to fill this gap.
Just like for single-agent settings, the goal is to sidestep the task of manually coding a game environment, and to instead input a game description into an LLM and prompt it for a model of the game as code for a standard library of game solvers, such as OpenSpiel \cite{LanctotEtAl2019OpenSpiel}.
We refer to \citeauthor{lehrach2025codeworldmodelsgeneral}'s CWMs for games as Game Code World Models, or \CWM{}s for short. 

Using LLMs for GameCWM-generation is a promising direction for streamlining their creation for novel multiagent environments.
Current approaches, however, face significant accessibility and scalability barriers.
For example, only frontier models (e.g., Gemini 2.5 Pro) have achieved acceptable success rates thus far \cite{lehrach2025codeworldmodelsgeneral}.
Furthermore, \citet{lehrach2025codeworldmodelsgeneral}, \citet{tang2024coderepairllmsgives}, \citet{dainese2024generatingcodeworldmodels}, and \citet{deng2025naturallanguageextensiveformgame} all rely on extensive test-time compute (i.e. computational effort exerted during inference) to search for, verify, and refine generated code.
More specifically, they use iterative-refinement loops, in which an LLM generates code, that is then checked---against a provided suite of unit tests%
---after which the LLM is called upon to try again, as needed. Our goal in this work is to tackle these barriers.
That is, our aim is to make the automated generation of syntactically and semantically correct GameCWMs more resource efficient.

The approach we take is a natural one.
We attempt to distill GameCWM generation into smaller models through post-training, i.e., refining a pre-trained base model through supervised and/or reinforcement learning \cite{jiang2026supervisedfinetuningversusreinforcement}. 
We hypothesize that reasoning about complex tasks, currently the burden of test-time compute, can be incorporated into post-training, whereby lightweight models would internalize 
capabilities that were previously available only in frontier LLMs.
\amy{has someone already demonstrated that this is possible in non-games? i.e., for Dainese's CWMs?} \tyrone{yeah i think dainese tests on non-games, like typical RL environments like cartpole}

\begin{figure}[h]
    \centering
    \includegraphics[width=\columnwidth]{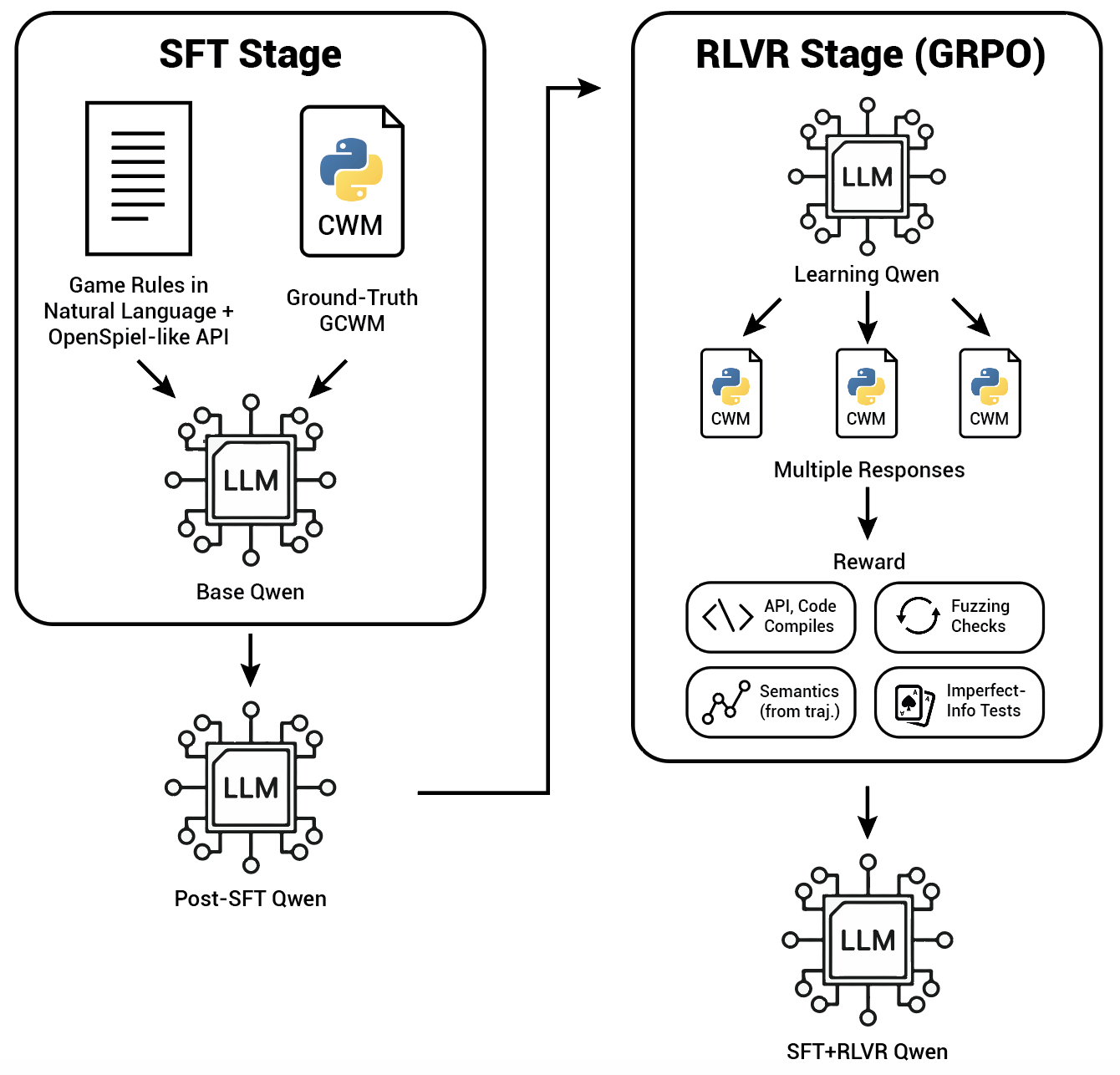}
    \caption{Our Distillation Pipeline. Unlike prior approaches that rely extensively on test-time compute, we aim to distill game-theoretic logic into the model weights through post-training. Like many LLM pipelines, ours involves two stages: Supervised Fine-Tuning (SFT) followed by \styrone{Group Relative Policy Optimization (GRPO)}{RLVR using GRPO}. To enable our approach, we define a dataset of prompt-code pairs for use during SFT, and a hierarchical verification framework for automatically evaluating generated code, which provides reward signals that reinforce the syntactic correctness and semantic validity of generated \CWM{}s during RLVR.}
    \label{fig:pipeline}
\end{figure}

We introduce a pipeline (see Figure ~\ref{fig:pipeline}) to post-train Qwen2.5-3B-Instruct \cite{qwen2.5} to generate GameCWMs for both perfect and imperfect information games.
Learning from a curated dataset of OpenSpiel-like implementations, our approach involves a Supervised Fine-Tuning (SFT) stage followed by a Reinforcement Learning with Verifiable Rewards (RLVR) stage \cite{lambert2025tulu3pushingfrontiers}.%
\footnote{Our code and dataset are available \href{https://github.com/tktserapio/internalizing-cwm-sft-grpo.git}{here}.}
In the latter, we use Group Relative Policy Optimization (GRPO) \cite{shao2024deepseekmath}, and we build an execution-based verifier that automatically assigns reward signals based on game-theoretic properties.



The specific contributions of this work are the following:
\begin{enumerate}
    \item \textbf{Dataset Curation:} We introduce a comprehensive dataset of extensive-form games implemented via an OpenSpiel-like API. The games were selected to test different axes of LLMs' abilities to generate \CWM{}s: mechanical state-tracking (primarily through perfect information games), imperfect information handling, and generalization to novel games.
    
    \item \textbf{Verification Framework:} Our framework validates LLM-generated \CWM{}s with respect to 1.~Python syntax; 2.~structural game-theoretic properties;
    and 3.~semantic game rule adherence. 
    To verify structural properties, we develop a property-based testing suite that evaluates \CWM{}s, without relying on
    oracle trajectories sourced from pre-existing game engines (e.g., OpenSpiel).
    To verify semantic adherence, we build an execution-based verifier that evaluates game trajectories produced by the CWM by comparing them to 
    trajectories produced directly from the game rules by a \emph{frontier\/} LLM. 
    This approach allows us to evaluate LLM-generated \CWM{}s even for novel games for which simulated game data is not available.
    
    \item \textbf{Distilling \CWM{} generation through Post-Training:} We show that post training (SFT + RLVR) can be used to distill game APIs and game logic into small open-weight LLMs, and thereby improve their generation of valid \CWM{}s in both perfect and imperfect information games.
\end{enumerate}

While we are able to demonstrate the potential of post-training for \CWM{} distillation in Qwen2.5-3B-Instruct (i.e., an LLM at the 3B parameter scale), as SFT and RLVR do improve its performance, there are limitations to our approach.
In particular, 
imperfect information games remain challenging.
We discuss these limitations and directions for future work in Section ~\ref{sec:conclusion}.

\section{Related Work}

Recent literature has explored the intersection of LLMs and game-theoretic environments. This section reviews previous work across: (1) the use of LLMs as strategic game-playing agents, (2) the emerging paradigm of leveraging LLMs to generate environment representations, and (3) the application of reinforcement learning to improve LLM code generation, which serves as the foundation for our post-training pipeline.

\subsection{LLMs as Game-Playing Agents} 

Perhaps the most natural approach is to treat the LLM as a player of the game, and simply prompt it to do some ``internal planning'' (i.e., to rollout game trajectories in text).
This approach, which assumes the LLM is equipped with a latent function over game continuations, has been explored for a diverse set of games, such as Chess \cite{feng2023chessgptbridgingpolicylearning} and Werewolf \cite{xu2024exploringlargelanguagemodels}. Diplomacy, however, relied on an external strategic reasoning engine to guide the LLM \cite{meta2022diplomacy}.

Although LLMs can achieve reasonable performance via internal planning in games that are well-represented in their training data, they face fundamental limitations when attempting to play complex \cite{schultz2025masteringboardgamesexternal} and novel games \cite{10.1609/aiide.v21i1.36804}. To ground LLMs to the underlying game structure, \citet{gemp2024steering} create a binding from natural language to the symbolic logic of games, which enables their game play to be externally steered by game solvers. 
\tyrone{arjun wrote this part so i'm not 100\% sure about the intent, but i tried rewriting it to make it clearer. hopefully it still captures the point @arjun!}

There is also increasing interest in investigating LLMs in multi-agent settings beyond the formal structure of games.
\citet{park2023generativeagentsinteractivesimulacra} introduced Generative Agents, an architecture that leverages LLMs to simulate believable social interactions in a ``sandbox" society, and \citet{zhou2024sotopiainteractiveevaluationsocial} introduced Sotopia, an open-ended environment designed to characterize the social intelligence of LLMs by simulating interactions in which agents must navigate complex, goal-driven social scenarios.
These directions further highlight the need to develop world models for multi-agent scenarios, as such models would facilitate a rigorous study of strategic social interaction.

\subsection{LLMs for Game and Environment Generation} 

Rather than using LLMs as players, an emerging direction aims to use LLMs to generate formal representations of game mechanics from natural language descriptions.
GG-Bench \cite{verma2025measuringgeneralintelligencegenerated} is such a pipeline.
It works by first generating games, then converting those games into Gymnasium-based \cite{towers2024gymnasium} environments, and finally training RL agents in those environments.
These RL agents can then be used to evaluate other game-playing agents, including LLMs themselves.



\citet{deng2025naturallanguageextensiveformgame} use LLMs to construct extensive-form game (EFG) tree representations in \path{pygambit} from natural language descriptions. Like us, they find that it is much more challenging to generate environments that correctly represent imperfect information games than perfect information games.
To tackle this problem, they introduce a two-stage framework: first, an LLM identifies information sets and partial tree structures; second, an LLM generates the EFG code.
The generation step utilizes an automated self-debugging loop where the generated code is executed, and then any Python error messages are fed back into the LLM for correction. \styrone{}{Manual checks are performed afterwards to ensure consistency with the game rules.}

\amy{also, ChatGPT tells me Deng, et al use a hierarchical approach, first attempting to extract the structure of the game, meaning the states, actions, and info sets, and then in setp 2, they construct the game tree. this seems like a good idea to me! did we try something like it?} \tyrone{i don't think we tried something like it unfortunately. but we definitely can - i was thinking of a more granular curriculum learning type thing where we simplify these complex cwms and increase difficulty but i wasn't able to implement it for this paper}


\citet{dainese2024generatingcodeworldmodels} explore the generation of Code World Models (CWMs) for model-based RL by leveraging LLMs to write executable Python representations of RL environments given natural language descriptions. To address the challenges of generating accurate transition and reward logic, they introduce Generate, Improve and Fix with Monte Carlo Tree Search (GIF-MCTS), an inference-time code generation strategy. GIF-MCTS guides the LLM to iteratively generate, self-debug, and refine candidate programs using feedback based on ground-truth environment trajectories. While highly effective, this methodology relies heavily on test-time compute to successfully search for and refine the generated code. Moreover, this work is limited to single-agent, perfect-information environments. \amy{plus, you need to have access to ground truth}

Our work is inspired by \citet{lehrach2025codeworldmodelsgeneral}, who used \CWM{}s for multi-agent general game playing, including imperfect-information game environments with partial observability. Similar to the approaches of \citet{dainese2024generatingcodeworldmodels} and \citet{deng2025naturallanguageextensiveformgame}, \citeauthor{lehrach2025codeworldmodelsgeneral}'s method relies on 
an iterative refinement loop, where a frontier LLM generates code, tests it using a suite of unit tests and ground-truth game trajectories, \amy{are these the same kind of ground-truth trajectories mentioned in the previous paragraph?}\tyrone{yep}and regenerates code upon failure.
While effective, these methodologies \cite{deng2025naturallanguageextensiveformgame, dainese2024generatingcodeworldmodels, lehrach2025codeworldmodelsgeneral} require substantial test-time compute and frontier models (e.g. GPT-4, Gemini 2.5 Pro) to achieve acceptable success rates.
We aim to distill \CWM{} generation into smaller models via post-training, ideally eliminating dependency on iterative refinement at test time and on frontier models.

\subsection{Reinforcement Learning for Code Generation}

Reinforcement learning (RL) has emerged as a critical post-training technique for improving code generation capabilities beyond what SFT alone can achieve \cite{le2022coderlmasteringcodegeneration}. More specifically, we use RLVR \cite{wang2025coevolvingllmcoderunit, jiang2026coderlimprovingcodegeneration} to generate a reward signal by verifying code output with test cases.
The credibility of this approach was demonstrated by DeepSeek-R1 \cite{Guo_2025}.
We also draw inspiration from hierarchical approaches such as StepCoder \cite{dou2024stepcoderimprovecodegeneration}, which breaks down complex coding challenges into sub-tasks, a technique we mimic in our proposed hierarchical verification framework.


Among the different approaches to RL
for LLM-based code generation, Group Relative Policy Optimization (GRPO) \cite{shao2024deepseekmath} has become a preferred choice, as compared to Proximal Policy Optimization (PPO) \cite{schulman2017proximalpolicyoptimizationalgorithms}, because verifiable rewards are clean signals, and as such are sufficient for computing the relative advantage of one candidate's reward as compared to that of the group.

\section{Background}
In this section, we define extensive-form games, which is the class of games of interest in this paper.
Extensive-form games are dynamic games that are represented as trees, which is an exponentially more compact representation than the classic normal form.

\subsection{Extensive-Form Games}

Extensive-form games (EFGs) are often used to model strategic interactions in multi-player sequential games \cite{10.5555/1483085}.
Such games are typically modeled by game trees, where each state of the game is represented by a node in the tree, and players' actions are represented by edges from one node to another.
Non-terminal nodes, where exactly one player moves, are decorated with the set of actions available to the player whose turn it is to move.
Terminal nodes, at which there are no further actions, are decorated with payoffs (e.g., win, lose, or draw in zero-sum games).
There is typically a chance player as well, who models stochasticity in the environment (e.g., card deals or dice rolls).

In some games, like Chess, all information is available to all players, while in others, like Poker, some information is hidden (e.g., cards).
The latter are called \textbf{imperfect-information} games.
In these games, the players' decision nodes are not individual states, but rather \textbf{information sets}, each of which is a set of states that the player to move cannot distinguish from one another.

We define a history \(h\) as a sequence of states and actions from the start of the game.
Reasoning about how to play at a given information set in imperfect-information games requires that the player be able to reconstruct the possible histories that could have led her to that information set.

\subsection{Game Code World Models (\CWM{}s)} 

We define a \CWM{} as an executable program generated by an LLM that implements the rules of a game environment as executable functions.

A \CWM{} provides functions that act as the engine for a game, including state transition dynamics, 
legal actions, and observations for each player.
We implement this interface in Python, using the following methods:

\begin{itemize}
    \item \texttt{apply\_action}: Returns the successor state after applying an action.
    
    \item \texttt{get\_legal\_actions}: Enumerates valid actions from a given state.
    
    \item \texttt{get\_current\_player}: Identifies which player acts (or indicates terminal/chance nodes).
    
    \item \texttt{get\_rewards}: Returns payoffs for each player at terminal states.
    
    \item \texttt{get\_observations}: Returns player-specific observations.
\end{itemize}

\noindent
Additionally, for imperfect-information games, we also require a \path{resample_history} function which, given a player's current information set, samples a complete history---i.e., a sequence of states and actions---consistent with her observations.

A key feature of \CWM{}s is their compatibility with 
game solvers. Once a syntactically correct \CWM{} is generated, Monte Carlo Tree Search (MCTS) \cite{coulom2006efficient} or Information-Set MCTS (ISMCTS) \cite{cowling2012information} can be applied immediately without additional training---assuming a semantically correct \CWM---enabling strong game play from a natural language specification alone.

\section{Methodology}
\label{sec:methodology}

Our approach comprises three components: (1) a curated dataset of 30 games spanning varying difficulty levels, (2) a hierarchical verification framework that evaluates the generated \CWM{}s, and (3) a two-stage post-training pipeline combining Supervised Fine-Tuning (SFT) with Group Relative Policy Optimization (GRPO).

\subsection{Dataset Construction}
\label{data_curriculum}

We construct a dataset of prompt-code pairs $(P_G, C_G)$, for every game \(G\), where the input prompts $P_G$ contain:

\begin{enumerate}
    \item The API specification outlining the required \CWM{} functions.
    
    \item The game description expressed in natural language.
    
    \item Explicit names of allowed action strings. By establishing naming conventions, we seek to avoid syntactic action-space mismatches, thereby enabling the verification framework to reward semantic adherence to the game rules.
    
    \item An example trajectory, or \textbf{scenario trace}, which is a natural language playthrough generated by a frontier LLM, simulating a human reporting their step-by-step experience traversing the environment.
\end{enumerate}

The target output $C_G$ contains the ground-truth Python code, including complete function implementations, imports, and type definitions and annotations.
We present our prompts in Appendix~\ref{pseudocode}; they are adapted from \citet{lehrach2025codeworldmodelsgeneral}.

We categorize games along the following axes, which guide the Supervised Fine-Tuning stage:

\begin{itemize}
    \item \textbf{Easy Games:} 
    Basic perfect information games with simple state spaces (e.g. Tic-Tac-Toe, Connect4).
    
    \item \textbf{Intermediate Games:} 
    More complex perfect information games (e.g. Y, Breakthrough) and slightly easier imperfect information games (e.g. Blackjack, Kuhn Poker).
    
    \item \textbf{Hard Games:} 
    More complex imperfect information games (e.g. Leduc Poker, Poker Hold'em).
    
    \item \textbf{Novel / Out-of-Distribution (OOD) Games:} 
    Perfect and imperfect information games that are not present in standard LLM training data (Hand of War, Quadranto, Generalized Tic-Tac-Toe, Generalized Chess \cite{lehrach2025codeworldmodelsgeneral}). 
    In this paper, we also introduce a new OOD game, Converge, created by the authors and Gemini 3 Pro.
\end{itemize}

To construct ground-truth implementations, we prompt Gemini 3 Pro to generate candidate ground-truth \CWM{}s based on game rules, and iteratively refine these models using a loop similar to that of \citet{lehrach2025codeworldmodelsgeneral}. 
Each implementation must pass a comprehensive test suite: for standard and pre-existing games, we adapt tests from OpenSpiel \cite{LanctotEtAl2019OpenSpiel} and also use our verification framework; for novel OOD games, we write custom tests, and again use our verification framework as well. 
This process is designed to ensure high-quality training data, while avoiding the computational cost of iterative refinement at test time.

\subsection{Verification Framework}
\label{hierarchical}

A core contribution of this work is a verification framework that evaluates \CWM{}s by combining property-based testing and lightweight game trajectory matching. We say that our form of trajectory matching is lightweight because rather than using pre-existing game engines to generate ground-truth game trajectories, we test against a small set of game traces created by the inputting the game rules to an LLM. This design offers two advantages: (1) it enables rigorous evaluation even of novel games for which ground-truth data do not exist, and (2) it provides a fast, sparse reward signal for RL training. The framework comprises four tiers. We provide the pseudocode for the implementation of each tier of the evaluation framework in Appendix~\ref{pseudocode}.

\paragraph{Tier 1: Static Analysis.} First, we verify syntactical and structural validity:

\begin{itemize}
    \item API Completeness: The class must implement all required methods listed in the prompt.
    
    \item Type signatures are repsected (e.g., \path{get_legal_actions} must return a list).
    
    \item All required libraries are imported.
    
    \item Code compiles cleanly.
\end{itemize}

\paragraph{Tier 2: Dynamics (Fuzzing).} We fuzz the generated \CWM{} with 100 random synthetic trajectories to ensure core physics engine invariants hold:

\begin{itemize}
    \item Determinism: Applying action $\action$ to state $\sstate$ must always yield the identical state $\sstate'$ (excluding chance nodes).
    
    \item Non-terminal Logic: If a state is non-terminal,
    \path{get_legal_actions} must return a non-empty list.
    
    \item Terminal Logic: If a state is terminal, 
    \path{get_legal_actions} must return an empty list.
    
    \item Crashes in Execution: The game must not raise exceptions (e.g., \path{KeyError}).
    
    \item State Immutability: \path{apply_action} must return a new state object.
\end{itemize}

\paragraph{Tier 3: Semantic Rule Adherence (LLM-generated Scenario Traces).} To verify that the generated engine adheres to the rules of the game, we evaluate the generated \CWM{}s against a small set of LLM traces we call \textbf{scenarios} derived \emph{a priori} from the game rules. Examples appear in Appendix \ref{scenarios}.

\begin{itemize}
    \item For each game, we prompt a frontier model with a game description, and ask it generate a sparse, representative suite of roughly five scenarios for that game, each of which is a history of gameplay.
    
    \item These scenarios test boundary conditions (e.g. player win states, maximum-turn draw limits, rejection of illegal actions) and typical gameplay.
    
    \item The reward signal for this tier is continuous: the candidate \CWM{} receives a score proportional to the fraction of scenarios it passes successfully.
\end{itemize}


\paragraph{Tier 4: Information Consistency.} 
In imperfect-information games, players do not have access to full state information. 
Following \citet{lehrach2025codeworldmodelsgeneral}, we evaluate the \texttt{resample\_history} method to ensure epistemic boundaries are maintained.

\begin{itemize}
    \item We simulate a random walk from the initial state to a valid state $\sstate$, extract player $\player$'s observation $\observation$ at $\sstate$.
    
    \item We then prompt \texttt{resample\_history} to generate a hypothetical history $\hhistory'$ to state $\sstate$.
    
    \item Finally, we execute $\hhistory'$ from the initial state, and assert whether the resulting observations, $\observation'$ and $\observation$, differ. 
    If they do, the model has failed.
\end{itemize}

Although this tier tests observation consistency across resampled histories, it does not verify against illegal information leakage (e.g. ensuring the underlying \texttt{state} dictionary the LLM created within the \CWM{} perfectly isolates hidden variables from the observation functions), which we will integrate in future work.

We also design reward computation to be gated: tier 1 must pass before dynamics are evaluated, and dynamics must score $\geq 0.5$ before tiers 3 and 4 contribute.
In addition, Qwen seems to recognize that \path{resample_history} is challenging to implement; it leaves stubs in the \CWM{} (i.e., returning an empty list).
As a result, we added a check for these stubs, and an accompanying penalty during RLVR to tier 4 of our verification framework. \amy{why tier 4 and not tier 3, which is semantics?} \tyrone{can reconsider for next version}



\subsection{Training Pipeline}

\paragraph{Supervised Fine-Tuning (SFT)} 
We initialize from Qwen2.5-3B-Instruct \cite{qwen2.5} and fine-tune on the $(\gameprompt, \gamecode)$ pairs. 
To support reproducibility, we specify our training hyperparameters and train/eval split in Appendix~\ref{training_hyperparams} and Appendix~\ref{games_dataset}, respectively.

\paragraph{RLVR via Group Relative Policy Optimization (GRPO)} 
Following a SFT phase, we apply GRPO as developed by \citet{shao2024deepseekmath} to further align the model with our hierarchical verification framework, thereby executing RLVR.

To generate candidate \CWM{}s during the RLVR stage (as well as during our evaluations), we prompt the model with $\gameprompt$ for a given game $\game$, and consider its zero-shot output.
For each generated \CWM, we then compute a composite reward:
\[
  r_i = \sum_{t=1}^{4} \weight_t \Score_t
\]  
where $\Score_t \in [0,1]$ is the verification score at tier $t$, meaning the proportion of tests that pass at tier $t$). 
We set $\weight_1=0.15$ (static), 
$\weight_2=0.25$ (dynamics), 
$\weight_3=0.3$ (semantics), and 
$\weight_4=0.3$ (information handling), weighing semantics and information handling more heavily to prioritize semantic game logic over mere API compliance. 



\section{Experiments}

We evaluate the extent to which post-training with SFT, GRPO, and SFT+GRPO enables the distillation of \CWM{} generation capabilities into small models, comparing the contributions of SFT and GRPO both independently and jointly.

\subsection{Experimental Setup}

\paragraph{Training.} We train four models built on top of Qwen2.5-3B-Instruct \cite{qwen2.5} on 23 of the 30 games in our data set. We use the remaining 7 as a held-out test set. The four models are:

\begin{itemize}
    \item \textbf{Base}: The original Qwen2.5-3B-Instruct before applying our post-training pipeline.
    
    \item \textbf{SFT}: Fine-tuned on our dataset using supervised learning using next-token prediction loss with LoRA \cite{hu2021loralowrankadaptationlarge}.
    
    \item \textbf{GRPO}: Base model trained directly with GRPO using our execution-based verification as a reward signal.
    
    \item \textbf{SFT+GRPO} (our full pipeline): SFT model further refined with GRPO.
\end{itemize}

\paragraph{Evaluation.} 
We evaluate each model on a set of 7 held-out games, which includes novel/OOD and non-novel (in-distribution) perfect- and imperfect-information games.%
\footnote{We consider the games introduced by \citet{lehrach2025codeworldmodelsgeneral} to be novel/OOD with Qwen2.5-3B-Instruct because the Gemini model they use was published after the release date of the Qwen2.5 models.}
The perfect-information games are Generalized Tic-Tac-Toe (OOD), Generalized Chess (OOD), Converge (OOD), Y; the imperfect-information games are Quadranto (OOD), Hand-of-War (OOD), and Gin Rummy. This setup is summarized in Table \ref{tab:eval_games_classification}.
For each, we generate 25 samples per game, and report average verification scores across the four verification tiers.%
\footnote{We set Qwen's temperature to be 0.3, to balance accuracy and flexibilty.} 

\begin{table}[htbp]
    \centering
    \caption{Categorization of held-out games based on information type and distribution.}
    \label{tab:eval_games_classification}
    \resizebox{\columnwidth}{!}{%
    \begin{tabular}{@{}lcc@{}}
        \toprule
        & \textbf{Perfect Information} & \textbf{Imperfect Information} \\ 
        \midrule
        \textbf{In-Distribution} & Y & Gin Rummy \\ 
        \addlinespace
        \textbf{Out-of-Distribution} & Gen. Tic-Tac-Toe, & Quadranto, \\
        \textbf{(OOD)} & Gen. Chess, Converge & Hand of War \\ 
        \bottomrule
    \end{tabular}
    }
    \label{tab:held_out_games}
\end{table}

\textit{Note:} A significant number of the outputs by the base Qwen model do not compile at first, simply because of missing imports.
To ensure that we are testing Qwen's ability to generate valid \CWM{}s, rather than its ability as a Python software engineer, we inject these missing imports via a preprocessing step. 

\subsection{Results}
\label{sec:results}

Table~\ref{tab:overall_ci} reports mean verification scores across all held-out games and all verification tiers for various confidence levels.
Table~\ref{tab:overall_mean} disaggregates these rates by tier, and Table~\ref{tab:model_results} further disaggregates by game.
Tier 1 has seven tests, Tier 2 has four tests, Tier 3 has five tests, and Tier 4 has four tests, with each tier's tests capturing the components discussed in Section~\ref{hierarchical}. 
A score of \(0.0\%\) indicates that no generated \CWM{} passed any test in that tier's verification suite. 
A few generated \CWM{}s of interest can be found in Appendix~\ref{code_samples}; the rest are published in our \href{https://github.com/tktserapio/internalizing-cwm-sft-grpo.git}{GitHub} repository.
\amy{presumably, the ones you include in the Appendix are interesting for some reason or another!!} 



Based on our experiments, several findings emerge.

\begin{table}[!t]
\centering
\caption{Overall mean verification score with 90\% and 95\% confidence intervals across held-out games. Intervals are computed over 25 generations per game. SFT+GRPO yields the strongest Qwen-based model. Note the statistically significant separation between it and the base model.}
\begin{tabular}{l c c c c}
\toprule
Model & Mean & 90\% CI & 95\% CI & N \\
\midrule
Base     & 47.3\% & [44.8, 49.8]\% & [44.3, 50.3]\% & 175 \\
SFT      & 49.0\% & [46.8, 51.2]\% & [46.4, 51.6]\% & 175 \\
GRPO     & 50.4\% & [48.6, 52.1]\% & [48.3, 52.4]\% & 175 \\
SFT+GRPO & 53.2\% & [51.3, 55.1]\% & [51.0, 55.4]\% & 175 \\
GPT-4o   & 66.7\% & [64.5, 69.0]\% & [64.1, 69.4]\% & 175 \\
\bottomrule
\end{tabular}
\label{tab:overall_ci}
\end{table}

\paragraph{Overall, SFT+GRPO post-training yields the strongest Qwen-based model}
As shown in Table~\ref{tab:overall_ci}, SFT+GRPO produces the most capable Qwen model in our study, demonstrating a statistically significant separation over the base model's performance. Despite these improvements, GPT-4o remains significantly stronger overall.


\begin{table}[!t]
      \centering
      \caption{Per-tier average verification scores across all held-out games. Values are averaged across 25 sample generations per game. Information is averaged only over imperfect-information games. SFT+GRPO yields the strongest performance in all categories among the post-trained models.}
      \label{tab:overall_mean}
      \resizebox{\columnwidth}{!}{
      \begin{tabular}{lcccc}
          \toprule
          \textbf{Model} & \textbf{Static} & \textbf{Fuzz} & \textbf{Semantics} & \textbf{Information} \\
          \midrule
          Base & 77.5\% & 70.2\% & 8.5\% & 3.3\% \\
          SFT & 78.0\% & 74.3\% & 9.4\% & 2.0\% \\
          GRPO & 83.3\% & 75.3\% & 8.4\% & 3.7\% \\
          SFT+GRPO & 86.3\% & 75.1\% & 14.4\% & 5.3\% \\
          GPT-4o & 98.4\% & 84.6\% & 33.8\% & 14.0\% \\
          \bottomrule
      \end{tabular}
      }
\end{table}

\paragraph{SFT improves syntactical and structural correctness.} As shown in Table~\ref{tab:overall_mean}, SFT can improve Syntax (Tier 1) and Fuzz (Tier 2) scores compared to the base model.
More specifically, in Table~\ref{tab:model_results}, we see that SFT increased the Fuzz score (or it is unchanged) in all games.
In addition, on most OOD games, SFT increases the Static score as compared to the base model. (An exception is Quadranto, but the decrease was small.) 

In Gin Rummy, however, an in-distribution game for the base model that we did not include in our SFT training set, SFT decreased performance as compared to the base model from 49.7\% to 31.4\%!
We inspected the generated code samples for SFT on Gin Rummy and uncovered several syntax errors, such as mixing list elements with a list comprehension \styrone{}{(e.g. \texttt{["Draw stock", "Draw upcard", "Action: " +
card for card in state["deck"]]}); see Appendix \ref{code_samples}).}
The usual explanation for SFT improving performance 
{is that it better aligns a model's output with specialized downstream outputs. In this case, however, SFT appears to have led to overfitting. The syntax errors demonstrate that the model blindly attempts to fit Gin Rummy's mechanics into the list-comprehension templates it learned during SFT, overriding the more syntactically valid code generation capabilities of the base model.}

\amy{since Gin Rummy was not in the training set, i think generalization could be a simpler/better explanation, except that the game is IN distribution, i.e., in the pre-training set?} \tyrone{similar to ur point about exploration, sft will make it mimic what is in the training set, and if gin rummy doesn't have the same nature as other games (e.g. it being a lot more complex), then sft won't work. but i guess im not an llm expert}

\begin{table}[t]
      \centering
      \caption{Aggregate verification scores across the four verification tiers. Values are averaged across 25 sample generations per game. Semantics are challenging for all models, including GPT-4o, as is information-handling in imperfect-information games.}
      \label{tab:model_results}
      \resizebox{\columnwidth}{!}{%
      \begin{tabular}{llcccc}
          \toprule
          & & \multicolumn{4}{c}{\textbf{Test Hierarchies}} \\
          \cmidrule(lr){3-6}
          \textbf{Game} & \textbf{Model} & \textbf{Static} & \textbf{Fuzz} & \textbf{Semantics} & \textbf{Information} \\
          \midrule
          \multicolumn{6}{l}{\textit{\textbf{Perfect Information Games}}} \\
          \midrule
          \multirow{5}{*}{\textbf{Gen. Tic-Tac-Toe (OOD)}}
            & Base & 90.3\% & 65.6\% & 17.7\% & N/A \\
            & SFT & 98.3\% & 68.0\% & 26.3\% & N/A \\
            & RLVR & 98.9\% & 65.6\% & 23.4\% & N/A \\
            & SFT+RLVR & 96.0\% & 62.4\% & 20.6\% & N/A \\
            & GPT-4o & 100.0\% & 100.0\% & 80.0\% & N/A \\
          \midrule

          \multirow{5}{*}{\textbf{Converge (OOD)}}
            & Base & 92.0\% & 78.4\% & 0.0\% & N/A \\
            & SFT & 97.7\% & 82.4\% & 0.0\% & N/A \\
            & RLVR & 96.6\% & 80.0\% & 1.6\% & N/A \\
            & SFT+RLVR & 91.4\% & 82.4\% & 0.0\% & N/A \\
            & GPT-4o & 96.0\% & 91.2\% & 52.0\% & N/A \\
          \midrule

          \multirow{5}{*}{\textbf{Gen. Chess (OOD)}}
            & Base & 61.1\% & 76.8\% & 1.6\% & N/A \\
            & SFT & 64.6\% & 76.8\% & 4.8\% & N/A \\
            & RLVR & 61.7\% & 73.6\% & 5.6\% & N/A \\
            & SFT+RLVR & 74.9\% & 76.8\% & 20.8\% & N/A \\
            & GPT-4o & 93.1\% & 80.0\% & 33.6\% & N/A \\
          \midrule

          \multirow{5}{*}{\textbf{Y}}
            & Base & 90.9\% & 75.2\% & 16.8\% & N/A \\
            & SFT & 92.0\% & 78.4\% & 15.2\% & N/A \\
            & RLVR & 92.6\% & 76.8\% & 15.2\% & N/A \\
            & SFT+RLVR & 97.7\% & 72.8\% & 34.4\% & N/A \\
            & GPT-4o & 100.0\% & 80.8\% & 19.2\% & N/A \\
          \midrule

          \multicolumn{6}{l}{\textit{\textbf{Imperfect Information Games}}} \\
          \midrule
          \multirow{5}{*}{\textbf{Quadranto (OOD)}}
            & Base & 84.6\% & 76.8\% & 16.8\% & 6.0\% \\
            & SFT & 82.3\% & 82.4\% & 12.8\% & 6.0\% \\
            & RLVR & 81.7\% & 80.0\% & 8.0\% & 5.0\% \\
            & SFT+RLVR & 86.9\% & 80.8\% & 15.2\% & 6.0\% \\
            & GPT-4o & 100.0\% & 88.0\% & 33.6\% & 18.0\% \\
          \midrule

          \multirow{5}{*}{\textbf{Hand of War (OOD)}}
            & Base & 73.7\% & 72.8\% & 5.6\% & 1.0\% \\
            & SFT & 79.4\% & 76.0\% & 6.4\% & 0.0\% \\
            & RLVR & 84.0\% & 76.8\% & 4.8\% & 1.0\% \\
            & SFT+RLVR & 89.7\% & 75.2\% & 9.6\% & 1.0\% \\
            & GPT-4o & 100.0\% & 72.0\% & 18.4\% & 19.0\% \\
          \midrule

          \multirow{5}{*}{\textbf{Gin Rummy}}
            & Base & 49.7\% & 45.6\% & 0.8\% & 3.0\% \\
            & SFT & 31.4\% & 56.0\% & 0.0\% & 0.0\% \\
            & RLVR & 67.4\% & 74.4\% & 0.0\% & 5.0\% \\
            & SFT+RLVR & 67.4\% & 75.2\% & 0.0\% & 9.0\% \\
            & GPT-4o & 100.0\% & 80.0\% & 0.0\% & 5.0\% \\
          \bottomrule
      \end{tabular}
      }
\end{table}

\paragraph{SFT and RLVR, alone and combined, improves Fuzz scores} As seen in Table~\ref{tab:overall_mean}, SFT, RLVR, and SFT+RLVR achieve Fuzz scores ahead of the base model.
This result suggests that SFT and RL both help the base model produce transition functions that are more robust under randomized rollouts, but the gains are smaller than those observed at the Static tier. In several games, generated \CWM{}s can compile and expose the required API, while still failing to preserve necessary properties for compatibility with standard game theory solvers like MCTS.

\tyrone{For example, a generated model might directly mutate the input state dictionary (e.g., \texttt{state["board"][action] = current_player}). This in-place modification modifies the search tree during MCTS, which often requires pure transition functions that preserve the parent state in order to explore alternate future branches}. \amy{what does this mean: ``immutability of states in state transitions''? can you give me an example of this failure case?} \tyrone{it's just that to be usable for MCTS it can't mutate the state dict itself since we want it to be recursive/pure - i'll put an example}

\paragraph{SFT combined with RLVR improves adherence to game rule semantics.} 
As seen in Table~\ref{tab:overall_mean}, the Semantics score of GPT-4o, a frontier model, is only 33.4\%. This finding indicates that translating natural language games into game trajectories that faithfully model a game via a \CWM{} is currently a difficult task.

Among all Qwen models, SFT+RLVR obtains the best Semantics score. This result suggests that RL is most useful when applied after SFT. This finding highlights the value of our two-stage post-training method: SFT teaches the model the expected API format and other structure, and RLVR then provides execution-based feedback that improves adherence to game rules.

Even RLVR alone generally improves Semantic adherence in all games (both perfect- and imperfect-information, and both in-distribution and OOD) on our held-out data set. Not in Quadranto, however.
\styrone{}{Upon inspecting generated Quadranto \CWM{}s from the RLVR model, we found that some implementations failed to update the current-player indicator to the terminal value after the game had ended.  It seems our current RLVR reward only allows the model to learn simplified transition patterns.}

\paragraph{Complex games, especially complex imperfect-information games, remain especially challenging.} Across Quadranto, Hand of War, and Gin Rummy, all Qwen models achieve low Information scores, as does GPT-4o. SFT+RLVR performs best among post-trained Qwen models with an average Information score of 5.3\%, compared with 14\% for GPT-4o.
This finding indicates that the logical complexity of these games exceeds the current capacity of both frontier models and our training pipeline.

\paragraph{LLMs explicitly indicate their limitations.} 
Although they were unable to fully generate the complex logic for \path{resample_history} most of the time, an interesting behavior observed in the models was the explicit comments indicating the complexity of implementation. For example, in several generated code samples, the SFT+GRPO Qwen model inserted comments stating that the resampling logic in \path{resample_history} was too difficult, opting for a simplified implementation instead (e.g., returning the last action in the history). An example appears in Appendix~\ref{code_samples}. 

Our findings suggest that the models possess a high-level understanding of the API requirements, but lack the requisite reasoning abilities to generate logic in a \CWM{} that fully adheres to the game rules (e.g., the \texttt{resample_history} function).

\paragraph{Summary} These results point to three main conclusions. First, SFT alone is not enough: it does not improve overall verification performance and can reduce performance on some tiers. Second, RLVR improves some structural and rollout properties, but direct RLVR from the base model is less effective than applying RLVR after SFT, which aligns with LLM post-training practices that use SFT before RLVR \cite{wei2025advancingmultimodalreasoningreinforcement}. Third, the full SFT+RLVR pipeline exhibits the strongest post-trained Qwen results, especially on syntactical correctness and aggregate semantic adherence, but both our models and GPT-4o and struggle with complex semantic and imperfect-information reasoning.

\paragraph{The limitations of SFT and RLVR}
Both SFT and RLVR are known to be limited by the capabilities of the base model. 
Work by \citet{zhou2023limaalignment} shows that SFT primarily aligns a model to a desired format or style; it does not teach it complex or novel reasoning skills.
\citet{le2022coderlmasteringcodegeneration} note that relying solely on a next-token prediction objective penalizes the model for exploring valid solutions that deviate from the ground-truth dataset, severely limiting SFT's reasoning capabilities on novel problems.

RLVR faces similar limitations. Recent work by Yue et al. \cite{yue2025limit-of-rlvr} suggests that the reasoning capabilities of RLVR-trained models are bounded by the base model's sampling distribution; that is, if a correct reasoning path does not exist within the base model's latent space with non-zero probability, RL cannot ``invent" it. Given that our base model is a lightweight 3B parameter model, it is possible that the complex "Theory of Mind" logic required for these games is simply absent from its pre-training distribution. 
Indeed, upon inspecting a few generated code samples for Gin Rummy, we found that the post-trained models just return a constant set of actions or simply the last action in the history, just as the base model does.


\section{Conclusion and Future Work}
\label{sec:conclusion}

Game theory serves as a framework in which to understand multi-agent interactions, from simple board games to high-stakes scenarios in politics, economics, and corporate competition \cite{10.5555/1483085}. 
En route to building AI agents that behave effectively and reliably in game-theoretic environments, some have argued that a first step is to endow them with game code world models (\CWM{}s), executable environments that enforce game rules and manage environment dynamics. While recent work has proposed the use of Large Language Models (LLMs) to construct \CWM{}s from natural language, these approaches still rely on computationally expensive test-time simulations and refinement loops, and multiple calls to frontier LLM APIs. In this work, we questioned whether this reliance on expensive compute could be avoided by shifting the computational burden to post-training a small model.

\tyrone{AI agents require executable environments in which they can plan, search, learn, and be evaluated. In game-theoretic settings, such environments must implement not only state transitions and rewards, but also legal actions, turn order, terminal conditions, player-specific observations, and, in imperfect-information games, hidden-state structure. While recent work has shown that LLMs can generate such environments from natural-language descriptions, existing approaches typically rely on frontier models and expensive inference-time refinement loops. In this work, we questioned whether this reliance on expensive compute at test-time could be avoided by shifting the computational burden to post-training a smaller model.}

To this end, we presented a pipeline for distilling \CWM-generation into lightweight LLMs through post-training, aiming to remove the need for frontier models and inference-time refinement loops. Our approach combines a curated game dataset, a hierarchical verification framework based on game-theoretic properties, and a two-stage SFT+RLVR training pipeline. Experiments demonstrate that SFT alone, RLVR alone, and our pipeline enables Qwen2.5-3B-Instruct to become more capable of generating valid \CWM{}s in both perfect and imperfect information games. The core of our approach is the use of execution-based verification as a reward signal. We use game-specific behavioral checks to evaluate structural and semantic properties of game environments, including state immutability, determinism, terminal-state behavior, rule adherence, and observation consistency. In this way, we can post-train \CWM{} generators, thereby reducing reliance on expensive frontier-model repair loops at inference time.

\textbf{Limitations.} Several limitations constrain our current results. First, our dataset, while diverse, is likely insufficient to cover the full complexity of game-theoretic reasoning, as seen in our enhanced models' performance on games like Hand of War, where all Qwen models achieve 0\% on imperfect information tests. Second, our evaluation only focuses on verification scores and does not evaluate downstream gameplay performance; a generated \CWM{} may pass our tests yet still contain subtle bugs affecting MCTS planning quality, which points to future work to complete the pipeline. Third, a limit to the semantics tier of our verifier is that our semantic tests rely on LLM-generated scenario traces. Even when produced by frontier models, these traces are not guaranteed to be correct or comprehensive.

\textbf{Future Work.} Several directions extend from this work:
\begin{itemize}

    \item \textbf{Improving the verification framework.} Our current verifier checks syntax, selected structural properties of game dynamics, semantic adherence on a small set of scenario traces, and observation consistency for imperfect-information games. However, it does not provide complete verification of the generated game. For example, for imperfect-information games, future verifiers should also check information-set correctness more rigorously, ensuring that observation functions only expose information available to the acting player.
    
    \item \textbf{Scaling data and model size.} Expanding the dataset to include more imperfect-information games and evaluating larger base models (e.g., 14B) may address the current failures on complex games and imperfect-information handling.
    
    \item \textbf{End-to-end gameplay evaluation.} Integrating generated \CWM{}s with MCTS and ISMCTS, and evaluating actual gameplay performance against reference implementations and/or human players could provide a more well-rounded, complete assessment of \CWM{} quality.
    
    \item \textbf{Iterative self-improvement.} Combining our trained model with iterative refinement over unit tests, trajectories, or even our constructed verification framework may recover the benefits of iterative refinement seen in the experiments of \citet{lehrach2025codeworldmodelsgeneral}, \citet{tang2024coderepairllmsgives}, and \citet{dainese2024generatingcodeworldmodels}. In particular, our finding of the models' "awareness" of their own limitations could guide an iterative refinement process.
\end{itemize}

Overall, our results suggest that \CWM{} generation can be partially distilled into lightweight LLMs through post-training. This work takes a step forward towards AI systems that can model and reason about complex strategic interactions in critical domains such as economic forecasting, corporate negotiations, and international relations, areas in which early-stage autonomous agents will surely benefit from incorporating language understanding.

\bibliographystyle{ACM-Reference-Format} 
\bibliography{sample}

\newpage 
\begin{appendices}
\section{Pseudocode}
\label{pseudocode}

\subsection{Perfect Information Game Generation Prompt}
\begin{lstlisting}
You are an expert Python programmer who is building the game of {game_name}.
Here is a description of the game:
{game_desc}

The goal is to implement Python functions with the following signatures and use the type definitions.
Your implementation MUST use module-level functions, NOT classes.
#START FUNCTION SIGNATURE
# Type definitions
Action = str
State = dict[str, Any]
PlayerObservation = dict[str, Any]

# Required Functions
def get_initial_state() -> State:
    """Returns the initial game state before any actions are taken."""
def apply_action(state: State, action: Action) -> State:
    """
    Returns the new state after an action has been taken.
    Ensure that the previous state is not mutated; always return a new state object.
    """
def get_current_player(state: State) -> int:
    """Returns current player (e.g. 0 or 1), or -4 for terminal state."""
def get_player_name(player_id: int) -> str:
    """Returns the name of the player."""
def get_rewards(state: State) -> list[float]:
    """Returns the rewards per player. May return non-zero values at non-terminal states if the game tracks running rewards (e.g., current scores or chip stacks); otherwise returns [0.0, 0.0] until meaningful reward information is available."""
def get_legal_actions(state: State) -> list[Action]:
    """Returns legal actions for current state. Empty list if terminal."""
def get_observations(state: State) -> list[PlayerObservation]:
    """Returns [player_0_obs, player_1_obs]. For perfect info games, both see the same state."""
#END FUNCTION SIGNATURE

Use the type definitions.
All actions returned by get_legal_actions and accepted by apply_action must use ONLY the action strings specified in the game description above. Do not invent new action formats.
Ensure to import all libraries used.
Do not leave placeholders.
Do not repeat the function signature.
Do write comments explaining what the code does.
Do use helper functions to reduce code duplication.
Do NOT wrap the functions in a class. Define them at the module level.
\end{lstlisting}
\subsection{Imperfect Information Game Generation Prompt}
\begin{lstlisting}
You are an expert Python programmer who is building the game of {game_name}.
Here is a description of the game:
{game_desc}

The goal is to implement Python functions with the following signatures and use the type definitions. 
Your implementation MUST use module-level functions, NOT classes.
#START FUNCTION SIGNATURE
# Type definitions
Action = str
State = dict[str, Any]
PlayerObservation = dict[str, Any]

# Required Functions
def get_initial_state() -> State:
    """Returns the initial game state before any actions are taken."""
def apply_action(state: State, action: Action) -> State:
    """
    Returns the new state after an action has been taken.
    Ensure that the previous state is not mutated; always return a new state object.
    """
def get_current_player(state: State) -> int:
    """Returns current player (e.g. 0 or 1), or -4 for terminal state."""
def get_player_name(player_id: int) -> str:
    """Returns the name of the player."""
def get_rewards(state: State) -> list[float]:
    """Returns the rewards per player. May return non-zero values at non-terminal states if the game tracks running rewards (e.g., current scores or chip stacks); otherwise returns [0.0, 0.0] until meaningful reward information is available."""
def get_legal_actions(state: State) -> list[Action]:
    """Returns legal actions for current state. Empty list if terminal."""
def get_observations(state: State) -> list[PlayerObservation]:
    """Returns [player_0_obs, player_1_obs]."""
def resample_history(obs_action_history: list[tuple[PlayerObservation, Action | None]], player_id: int) -> list[Action]:
    """
    Stochastically sample a valid sequence of actions (including 'chance' outcomes) that explains the current observations.
    CRITICAL: The returned list must be a complete trajectory that can be replayed starting EXACTLY from get_initial_state().
    """
#END FUNCTION SIGNATURE

Use the type definitions.
All actions returned by get_legal_actions and accepted by apply_action must use ONLY the action strings specified in the game description above. Do not invent new action formats.
Ensure to import all libraries used.
Do not leave placeholders.
Do not repeat the function signature.
Do write comments explaining what the code does.
Do use helper functions to reduce code duplication.
Do NOT wrap the functions in a class. Define them at the module level.
\end{lstlisting}

\subsection{Tiered Evaluation}
\begin{lstlisting}[mathescape=true]
# Tier weights (renormalized if information tier is skipped)                
TIER_WEIGHTS = {"static": 0.15, "dynamics": 0.25, "information": 0.30, "scenarios": 0.30}                 
def evaluate(code, game) -> float:
    module = load(code)

    # N is the number of games to fuzz on
    tiers = [static(module), dynamics(module, N=100), scenarios(module, game)]
    if game in IMPERFECT_INFO_GAMES:
        tiers.append(information(module, N=100))
    
    return mean(tier.score for tier in tiers)

def static(module) -> TierResult:
    yield syntax_ok # code compiled without error       
    
    yield all(hasattr(module, f) for f in REQUIRED_API) # if the 6 functions present
    
    if api_incomplete: yield [False] * 5; return
    
    state = module.get_initial_state()
    yield isinstance(state, dict)
    
    if state is None: yield [False] * 4; return
    yield returns list[str]:  module.get_legal_actions(state)
    
    yield returns list[float]: module.get_rewards(state)
    yield returns list: module.get_observations(state)
    yield returns int: module.get_current_player(state) 
    
def dynamics(module, N) -> TierResult:
    # 4 binary checks - pass iff property holds across all N trajectories
    for _ in range(N):
        track no_crash # no exception
        track immutable # deepcopy(s) == s after apply_action(s,a)
        track deterministic
        track terminal_empty
        
def information(module, N) -> TierResult:
    if not hasattr(module, "resample_history"): yield [False] * 4; return
    
    for _ in range(N): # pid is player id (0 or 1)
        history = [(obs[pid], action) for each of pid's turns]
        
        resampled = module.resample_history(history, pid)
        
        replay resampled from get_initial_state(), at each pid turn:
            yield resample_legal: action in get_legal_actions(state)
            
            yield obs_reconstruction: get_observations(state)[pid] == recorded_obs
            
            yield action_consistency: resampled_action == recorded_action
            
            yield resample_complete: all history entries consumed exactly
            
def scenarios(module, game) -> TierResult:
    for s in load_scenarios(game):
        state = replay(s.actions)
        yield terminal flag, current_player, rewards_sign, winner all match s.checks
\end{lstlisting}

\subsection{GRPO Reward}
\begin{lstlisting}
def compute_reward(code, game, N=20, timeout=60) -> float:
    with sigalrm(timeout): return _reward(code, game, N)  # 0.0 on timeout (infinite loop)

def _reward(code, game, N) -> float:
    W = renormalized_weights(game) # no information tests for perfect info games
    module, err = load(code)
    
    if err: return 0.0

    # Tier 1
    s_static, can_continue = check_static(module)
    
    if not can_continue:
        return s_static * W["static"]
        
    # Tier 2 - Dynamics
    s_dyn = check_dynamics(module, N)                   
    
    if s_dyn < 0.5:
        return s_static * W["static"] + s_dyn * W["dynamics"]  
    
    total = s_static * W["static"] + s_dyn * W["dynamics"]

    # Tier 3 - Scenarios  (fraction of golden scenarios passed)
    
    total += check_scenarios(module, game) * W["scenarios"]
    
    # Tier 4 - Information  
    
    if "information" in W:
        if is_stub(module.resample_history): return total
        
        total += check_information(module, N) * W["information"]

    return total
\end{lstlisting}

\section{Scenario Samples}
\label{scenarios}
\subsection{Leduc Poker}

\begin{table}[h]
\caption{Evaluation scenarios for Leduc Poker.}
\label{tab:scenarios-leduc}
\resizebox{\columnwidth}{!}{%
\begin{tabular}{p{4.8cm} p{6.0cm} l}
\toprule
\textbf{Scenario} & \textbf{Action sequence} & \textbf{Outcome} \\
\midrule
P0 folds preflop
  & \texttt{deal:K, deal:Q, Fold}
  & P1 wins \\[2pt]
P1 folds after P0 raises
  & \texttt{deal:K, deal:Q, Raise, Fold}
  & P0 wins \\[2pt]
Showdown: K beats J
  & \texttt{deal:K, deal:J, Call, Call, deal:Q, Call, Call}
  & P0 wins \\[2pt]
Pair beats non-pair (P0: K + public K)
  & \texttt{deal:K, deal:J, Call, Call, deal:K, Call, Call}
  & P0 wins \\[2pt]
P0 folds postflop facing P1 raise
  & \texttt{deal:K, deal:Q, Call, Call, deal:J, Call, Raise, Fold}
  & P1 wins \\[2pt]
Non-terminal after preflop deal
  & \texttt{deal:K, deal:Q}
  & P0 to act \\
\bottomrule
\end{tabular}
}
\end{table}

\subsection{Generalized Tic-Tac-Toe}
\begin{table}[h]
\caption{Evaluation scenarios for Generalized Tic-Tac-Toe.}
\label{tab:scenarios-gen-ttt}
\resizebox{\columnwidth}{!}{%
\begin{tabular}{p{4.8cm} p{6.5cm} l}
\toprule
\textbf{Scenario} & \textbf{Action sequence} & \textbf{Outcome} \\
\midrule
P0 wins: top row, cols 0--3
  & \texttt{0,0 $\to$ 1,0 $\to$ 0,1 $\to$ 1,1 $\to$ 0,2 $\to$ 1,2 $\to$ 0,3}
  & P0 wins \\[2pt]
P0 wins: column 0, rows 0--3
  & \texttt{0,0 $\to$ 0,1 $\to$ 1,0 $\to$ 0,2 $\to$ 2,0 $\to$ 0,3 $\to$ 3,0}
  & P0 wins \\[2pt]
P0 wins: main diagonal
  & \texttt{0,0 $\to$ 0,1 $\to$ 1,1 $\to$ 0,2 $\to$ 2,2 $\to$ 0,3 $\to$ 3,3}
  & P0 wins \\[2pt]
P1 wins: second row, cols 0--3
  & \texttt{0,0 $\to$ 1,0 $\to$ 2,0 $\to$ 1,1 $\to$ 3,0 $\to$ 1,2 $\to$ 4,0 $\to$ 1,3}
  & P1 wins \\[2pt]
P1 wins: column 5, rows 2--5
  & \texttt{0,0 $\to$ 2,5 $\to$ 0,1 $\to$ 3,5 $\to$ 0,2 $\to$ 4,5 $\to$ 1,0 $\to$ 5,5}
  & P1 wins \\[2pt]
Three-in-a-row is not terminal
  & \texttt{0,0 $\to$ 1,0 $\to$ 0,1 $\to$ 1,1 $\to$ 0,2}
  & P1 to act \\[2pt]
Non-terminal after first move
  & \texttt{3,3}
  & P1 to act \\
\bottomrule
\end{tabular}
}
\end{table}

\section{Training Hyperparameters}
\label{training_hyperparams}
\begin{table}[H]
\centering
\caption{Training hyperparameters for SFT and GRPO. If not indicated, defaults were used.}
\label{tab:hyperparams}
\resizebox{\columnwidth}{!}{%
\begin{tabular}{lccc}
\toprule
\textbf{Hyperparameter} & \textbf{SFT} & \textbf{GRPO} & \textbf{SFT+GRPO} \\
\midrule
\multicolumn{3}{l}{\textit{General}} \\
\midrule
Base Model & Qwen2.5-3B-Instruct & Qwen2.5-3B-Instruct & Post-SFT Model \\
Precision & bfloat16 & bfloat16 & bfloat16 \\
Epochs & 2 & 2 & 2 \\
Learning Rate & $1 \times 10^{-5}$ & $2 \times 10^{-5}$ & $2 \times 10^{-5}$ \\
LR Scheduler & Cosine & -- & -- \\
Warmup Ratio & 0.03 & -- & -- \\
Batch Size & 1 & 1 & 1 \\
Gradient Accumulation & 4 & 4 & 4 \\
Effective Batch Size & 4 & 4 & 4 \\
Optimizer & AdamW & AdamW 8-bit & AdamW 8-bit \\
Max Prompt Length & 4096 & 4096 & 4096 \\
Max Completion Length & 4096 & 4096 & 4096 \\
Gradient Checkpointing & -- & \checkmark & \checkmark \\
\midrule
\multicolumn{3}{l}{\textit{LoRA Configuration}} \\
\midrule
LoRA Rank ($r$) & 16 & 16 & 16 \\
LoRA Alpha ($\alpha$) & 32 & 64 & 64 \\
LoRA Dropout & 0.05 & 0.05 & 0.05 \\
Target Modules & \texttt{q,k,v,o\_proj} & \texttt{q,k,v,o,gate,up,down\_proj} & \texttt{q,k,v,o,gate,up,down\_proj} \\
\midrule
\multicolumn{3}{l}{\textit{GRPO-Specific}} \\
\midrule
Num Generations ($G$) & -- & 8 & 8 \\
Generation Batch Size & -- & 8 & 8 \\
KL Penalty ($\beta$) & -- & 0.1 & 0.1 \\
Max Gradient Norm & -- & 0.1 & 0.1 \\
Temperature & -- & 0.6 & 0.6 \\
\bottomrule
\end{tabular}
}
\end{table}

\section{Dataset Information}
\label{games_dataset}
\begin{table}[H]
\centering
\caption{Games used in our dataset, categorized by train/eval usage, if it's Existing/OOD, and information type}
\label{tab:games}
\resizebox{\columnwidth}{!}{%
\begin{tabular}{lcc}
\toprule
\textbf{Game} & \textbf{Existing/OOD} & \textbf{Information Type} \\
\midrule
\multicolumn{3}{l}{\textit{Training Games}} \\
\midrule
Tic-Tac-Toe & Existing & Perfect \\
Connect4 & Existing & Perfect\\
Pig & Existing & Perfect \\
Dots and Boxes & Existing & Perfect\\

Go & Existing & Perfect\\
Pentago & Existing & Perfect\\
Breakthrough & Existing & Perfect\\
Matching Pennies & Existing & Imperfect\\
Prisoner's Dilemma & Existing & Imperfect\\
Blackjack & Existing & Imperfect\\
Kuhn Poker & Existing & Imperfect\\
First Price Auction & Existing & Imperfect\\

Checkers & Existing & Perfect\\
Chess & Existing & Perfect\\
Liar's Dice & Existing & Imperfect\\
Goofspiel & Existing & Imperfect\\
Battleship & Existing & Imperfect\\
Hanabi & Existing & Imperfect\\

Leduc Poker & Existing & Imperfect\\
Backgammon & Existing & Perfect\\
Havannah & Existing & Perfect\\
Negotiation & Existing & Imperfect \\
Poker Hold'em & Existing & Imperfect \\
\midrule
\multicolumn{3}{l}{\textit{Evaluation Games (Held Out)}} \\
\midrule
Y & Existing & Perfect \\
Generalized Tic-Tac-Toe & OOD & Perfect\\
Generalized Chess & OOD & Perfect\\
Converge & OOD & Perfect \\
Gin Rummy & Existing & Imperfect \\
Hand of War & OOD & Imperfect\\
Quadranto & OOD & Imperfect\\
\bottomrule
\end{tabular}
}
\end{table}

\section{Ablation Studies}
\label{ablations}
\subsection{Ablation on Scenarios in Reward Function for GRPO}
Table \ref{tab:ablation_model_results} shows the results of ablating scenarios from the reward function in GRPO. We opted to report results including scenarios in the main text due to higher overall results on each category.
\begin{table}[h]
    \centering
    \caption{Aggregate pass rates (\%) across the four verification tiers (Static, Fuzzing, Semantics, Information) for only SFT+RLVR with and without scenarios. 
    Values represent average pass rates over 5 samples at temperature 0.3.}
    \label{tab:ablation_model_results}
    \resizebox{\columnwidth}{!}{%
    \begin{tabular}{llcccc}
        \toprule
        & & \multicolumn{4}{c}{\textbf{Test Hierarchies}} \\ 
        \cmidrule(lr){3-6}
        \textbf{Game} & \textbf{Model} & \textbf{Static} & \textbf{Fuzz} & \textbf{Semantics} & \textbf{Information} \\
        \midrule
        \multicolumn{6}{l}{\textit{\textbf{Perfect Information Games}}} \\
        \midrule
        \multirow{2}{*}{\textbf{Gen. Tic-Tac-Toe (OOD)}} 
          & SFT+RLVR (w/o scenarios) & 82.9\% & 75.0\% & 14.3\% & N/A \\
          & SFT+RLVR & 100.0\% & 75.0\% & 22.9\% & N/A \\
        \bottomrule

        \multirow{2}{*}{\textbf{Converge (OOD)}} 
          & SFT+RLVR (w/o scenarios) & 88.6\% & 80.0\% & 0.0\% & N/A \\
          & SFT+RLVR & 88.6\% & 75.0\% & 8.0\% & N/A \\
        \bottomrule

        \multirow{2}{*}{\textbf{Gen. Chess (OOD)}} 
          & SFT+RLVR (w/o scenarios) & 80.0\% & 75.0\% & 8.0\% & N/A \\
          & SFT+RLVR & 80.0\% & 90.0\% & 12.0\% & N/A \\
        \bottomrule

        \multirow{2}{*}{\textbf{Y}} 
          & SFT+RLVR (w/o scenarios) & 97.1\% & 80.0\% & 24.0\% & N/A \\
          & SFT+RLVR & 97.1\% & 80.0\% & 32.0\% & N/A \\
        \bottomrule
        \multicolumn{6}{l}{\textit{\textbf{Imperfect Information Games}}} \\
        \midrule
        \multirow{2}{*}{\textbf{Quadranto (OOD)}} 
          & SFT+RLVR (w/o scenarios) & 94.3\% & 70.0\% & 16.0\% & 0.0\% \\
          & SFT+RLVR & 85.7\% & 85.0\% & 12.0\% & 10.0\% \\
        \bottomrule

        \multirow{2}{*}{\textbf{Hand of War (OOD)}} 
          & SFT+RLVR (w/o scenarios) & 80.0\% & 60.0\% & 12.0\% & 0.0\% \\
          & SFT+RLVR & 77.1\% & 75.0\% & 4.0\% & 0.0\% \\
        \bottomrule

        \multirow{2}{*}{\textbf{Gin Rummy}} 
          & SFT+RLVR (w/o scenarios) & 68.6\%  & 75.0\% & 0.0\% & 0.0\% \\
          & SFT+RLVR & 82.9\% & 75.0\% & 0.0\% & 0.0\% \\
        \bottomrule  
        \multicolumn{6}{l}{\textit{\textbf{Aggregate Results}}} \\
        \bottomrule
        \multirow{2}{*}{\textbf{}} 
          & SFT+RLVR (w/o scenarios) & 78.0\%  & 75.7\% & 3.3\% & 0.0\% \\
          & SFT+RLVR & 87.3\% & 79.3\% & 13.0\% & 3.3\% \\
        \bottomrule  
    \end{tabular}
    }
\end{table}

\section{Samples of Generated Code}
\label{code_samples}
\subsection{SFT Qwen (Gin Rummy)}
\begin{lstlisting}[language=Python]
import copy
import random
from copy import deepcopy
from typing import List, Dict, Any, Optional, Tuple
from collections import defaultdict, Counter

from typing import Any, List, Dict, Tuple

Action = str
State = Dict[str, Any]
PlayerObservation = Dict[str, Any]

def initialize_deck() -> List[int]:
    """Initialize a shuffled deck of cards."""
    suits = ['S', 'C', 'D', 'H']
    ranks = [str(i) for i in range(2, 11)] + ['J', 'Q', 'K', 'A']
    deck = [(rank, suit) for suit in suits for rank in ranks]
    import random
    random.shuffle(deck)
    return deck

def split_deck(deck: List[Tuple[str, str]]) -> Tuple[List[Tuple[str, str]], List[Tuple[str, str]]]:
    """Split the deck into stock and upcard."""
    return deck[:-2], deck[-2:]

def get_initial_state() -> State:
    """Returns the initial game state before any actions are taken."""
    deck = initialize_deck()
    stock, upcard = split_deck(deck)
    return {
        'deck': deck,
        'stock': stock,
        'upcard': upcard,
        'current_player': 0,
        'knocked': False,
        'knock_card': 10,
        'deadwood': {'player_0': 0, 'player_1': 0},
        'melds': {'player_0': [], 'player_1': []}
    }

def apply_action(state: State, action: Action) -> State:
    """Returns the new state after an action has been taken."""
    new_state = state.copy()
    if action == 'Draw stock':
        new_state['stock'], new_state['upcard'] = split_deck(new_state['deck'])
        new_state['current_player'] = (new_state['current_player'] + 1) % 2
    elif action == 'Draw upcard':
        new_state['upcard'] = new_state['stock'].pop()
        new_state['current_player'] = (new_state['current_player'] + 1) % 2
    elif action.startswith('Action: '):
        card = action.split(': ')[1]
        new_state['deck'].remove((card, new_state['upcard'][0]))
        new_state['stock'].append((card, new_state['upcard'][0]))
        new_state['current_player'] = (new_state['current_player'] + 1) % 2
    elif action == 'Knock':
        new_state['knocked'] = True
        new_state['knock_card'] = 10
        new_state['knock_phase'] = 'Declare Melds'
    elif action == 'Done':
        new_state['knock_phase'] = 'Lay Off'
    elif action == 'Pass':
        new_state['knock_phase'] = 'Pass'
    return new_state

def get_current_player(state: State) -> int:
    """Returns current player (e.g. 0 or 1), or -4 for terminal state."""
    return state['current_player']

def get_player_name(player_id: int) -> str:
    """Returns the name of the player."""
    return f'Player {player_id}'

def get_rewards(state: State) -> List[float]:
    """Returns the rewards per player. May return non-zero values at non-terminal states if the game tracks running rewards (e.g., current scores or chip stacks); otherwise returns [0.0, 0.0] until meaningful reward information is available."""
    return [0.0, 0.0]

def get_legal_actions(state: State) -> List[Action]:
    """Returns legal actions for current state. Empty list if terminal."""
    current_player = get_current_player(state)
    if state['knocked']:
        if state['knock_phase'] == 'Declare Melds':
            return ['Action: ' + card for card in state['deck']]
        elif state['knock_phase'] == 'Lay Off':
            return []
        else:  # 'Pass' or 'Done'
            return []
    else:
        if state['current_player'] == current_player:
            if state['knock_phase'] == 'Draw':
                return ['Draw stock', 'Draw upcard']
            else:
                return ['Draw stock', 'Draw upcard', 'Action: ' + card for card in state['deck']]
        else:
            return []

def get_observations(state: State) -> List[PlayerObservation]:
    """Returns [player_0_obs, player_1_obs]."""
    player_0_obs = {
        'deck': state['deck'],
        'stock': state['stock'],
        'upcard': state['upcard'],
        'deadwood': state['deadwood']['player_0'],
        'melds': state['melds']['player_0'],
        'knock_phase': state['knock_phase']
    }
    player_1_obs = {
        'deck': state['deck'],
        'stock': state['stock'],
        'upcard': state['upcard'],
        'deadwood': state['deadwood']['player_1'],
        'melds': state['melds']['player_1'],
        'knock_phase': state['knock_phase']
    }
    return [player_0_obs, player_1_obs]

def resample_history(obs_action_history: List[Tuple[PlayerObservation, Action | None]], player_id: int) -> List[Action]:
    """Stochastically sample a valid sequence of actions (including 'chance' outcomes) that explains the current observations.
    CRITICAL: The returned list must be a complete trajectory that can be replayed starting EXACTLY from get_initial_state().
    """
    # Placeholder for resampling logic
    # For simplicity, we just return the most recent action
    return [action for _, action in obs_action_history]
\end{lstlisting}

\subsection{SFT+RLVR Qwen (Quadranto)}
\begin{lstlisting}[language=Python]
import copy
import random
from copy import deepcopy
from typing import List, Dict, Any, Optional, Tuple
from collections import defaultdict, Counter

import random

# Type definitions
Action = str
State = dict[str, Any]
PlayerObservation = dict[str, Any]

# Helper function to generate a random initial state
def get_random_initial_location():
    locations = [(0, 0), (0, 1), (1, 0), (1, 1), (0, 2), (0, 3), (1, 2), (1, 3), (2, 0), (2, 1), (3, 0), (3, 1), (2, 2), (2, 3), (3, 2), (3, 3)]
    return random.choice(locations)

# Required Functions
def get_initial_state() -> State:
    """Returns the initial game state before any actions are taken."""
    # Initial state: Player 0 in Q1, Player 1 in Q4
    p0_loc = get_random_initial_location()
    p1_loc = get_random_initial_location()
    while p0_loc == p1_loc:
        p1_loc = get_random_initial_location()

    # Mapping of quadrant to coordinates
    quadrant_mapping = {
        "Q1": [(0, 0), (0, 1), (1, 0), (1, 1)],
        "Q2": [(0, 2), (0, 3), (1, 2), (1, 3)],
        "Q3": [(2, 0), (2, 1), (3, 0), (3, 1)],
        "Q4": [(2, 2), (2, 3), (3, 2), (3, 3)]
    }

    # Determine initial quadrants
    q0 = next((q for q, coords in quadrant_mapping.items() if p0_loc in coords), None)
    q1 = next((q for q, coords in quadrant_mapping.items() if p1_loc in coords), None)

    return {
        "p0_loc": p0_loc,
        "p1_loc": p1_loc,
        "current_player": 0,  # Player 0 starts first
        "q0": q0,
        "q1": q1
    }

def apply_action(state: State, action: Action) -> State:
    """
    Returns the new state after an action has been taken.
    Ensure that the previous state is not mutated; always return a new state object.
    """
    new_state = state.copy()

    if action == "place_p0:<row>,<col>":
        row, col = map(int, action.split(":")[1].split(","))
        new_state["p0_loc"] = (row, col)
        new_state["q0"] = next(q for q, coords in quadrant_mapping.items() if (row, col) in coords)
    elif action == "place_p1:<row>,<col>":
        row, col = map(int, action.split(":")[1].split(","))
        new_state["p1_loc"] = (row, col)
        new_state["q1"] = next(q for q, coords in quadrant_mapping.items() if (row, col) in coords)
    else:
        # Movement actions
        row, col = new_state["p0_loc"]
        if action == "Up":
            new_state["p0_loc"] = (row - 1, col)
        elif action == "Down":
            new_state["p0_loc"] = (row + 1, col)
        elif action == "Left":
            new_state["p0_loc"] = (row, col - 1)
        elif action == "Right":
            new_state["p0_loc"] = (row, col + 1)
        elif action == "Stay":
            pass

    # Check if the game ended
    if abs(new_state["p0_loc"][0] - new_state["p1_loc"][0]) <= 1 and abs(new_state["p0_loc"][1] - new_state["p1_loc"][1]) <= 1:
        new_state["game_over"] = True
        new_state["winner"] = 1 if new_state["p0_loc"] == new_state["p1_loc"] else 0
        new_state["reward"] = [new_state["winner"], -new_state["winner"]]
    else:
        new_state["game_over"] = False
        new_state["reward"] = [0.0, 0.0]

    return new_state

def get_current_player(state: State) -> int:
    """Returns current player (e.g. 0 or 1), or -4 for terminal state."""
    return state.get("current_player", -4)

def get_player_name(player_id: int) -> str:
    """Returns the name of the player."""
    return f"Player {player_id}"

def get_rewards(state: State) -> list[float]:
    """Returns the rewards per player. May return non-zero values at non-terminal states if the game tracks running rewards (e.g., current scores or chip stacks); otherwise returns [0.0, 0.0] until meaningful reward information is available."""
    return state["reward"]

def get_legal_actions(state: State) -> list[Action]:
    """Returns legal actions for current state. Empty list if terminal."""
    if state["game_over"]:
        return []
    current_player = state["current_player"]
    legal_actions = []

    if current_player == 0:
        legal_actions.append("place_p0:<row>,<col>")
        legal_actions.append("Up")
        legal_actions.append("Down")
        legal_actions.append("Left")
        legal_actions.append("Right")
        legal_actions.append("Stay")
    else:
        legal_actions.append("place_p1:<row>,<col>")
        legal_actions.append("Up")
        legal_actions.append("Down")
        legal_actions.append("Left")
        legal_actions.append("Right")
        legal_actions.append("Stay")

    return legal_actions

def get_observations(state: State) -> list[PlayerObservation]:
    """Returns [player_0_obs, player_1_obs]."""
    p0_loc = state["p0_loc"]
    p1_loc = state["p1_loc"]
    p0_quadrant = state["q0"]
    p1_quadrant = state["q1"]

    p0_obs = {
        "My Loc": p0_loc,
        "Opponent Quadrant": p1_quadrant
    }
    p1_obs = {
        "My Loc": p1_loc,
        "Opponent Quadrant": p0_quadrant
    }

    return [p0_obs, p1_obs]

def resample_history(obs_action_history: list[tuple[PlayerObservation, Action | None]], player_id: int) -> list[Action]:
    """
    Stochastically sample a valid sequence of actions (including 'chance' outcomes) that explains the current observations.
    CRITICAL: The returned list must be a complete trajectory that can be replayed starting EXACTLY from get_initial_state().
    """
    # For simplicity, we'll just return a deterministic sequence of actions based on the current state
    if player_id == 0:
        return ["place_p0:0,0", "Up", "Down", "Left", "Right", "Stay"]
    else:
        return ["place_p1:3,3", "Up", "Down", "Left", "Right", "Stay"]
\end{lstlisting}

\subsection{RLVR Qwen (Quadranto)}
\begin{lstlisting}
import copy
import random
from copy import deepcopy
from typing import List, Dict, Any, Optional, Tuple
from collections import defaultdict, Counter

from typing import *
import random

# Type definitions
Action = str
State = dict[str, Any]
PlayerObservation = dict[str, Any]

# Helper function to generate a random position within a quadrant
def random_position_in_quadrant(quadrant: str) -> tuple[int, int]:
    rows, cols = (0, 1), (0, 3), (2, 1), (2, 3)
    row, col = rows[quadrant[0]], cols[quadrant[1]]
    return random.choice([row, row + 1]), random.choice([col, col + 1])

# Required Functions
def get_initial_state() -> State:
    # Initial positions
    p0_pos = random_position_in_quadrant('Q1')
    p1_pos = random_position_in_quadrant('Q4')
    
    # Initial state
    return {
        'p0_pos': p0_pos,
        'p1_pos': p1_pos,
        'turn_count': 0,
        'current_player': 0,
        'game_over': False
    }

def apply_action(state: State, action: Action) -> State:
    new_state = state.copy()
    p0_pos, p1_pos = new_state['p0_pos'], new_state['p1_pos']
    
    if action == 'place_p0:<row>,<col>':
        new_state['p0_pos'] = eval(action.split(':')[1])
    elif action == 'place_p1:<row>,<col>':
        new_state['p1_pos'] = eval(action.split(':')[1])
    else:
        if action == 'Up':
            new_state['p0_pos'] = (max(p0_pos[0] - 1, 0), p0_pos[1])
            new_state['p1_pos'] = (min(p1_pos[0] + 1, 3), p1_pos[1])
        elif action == 'Down':
            new_state['p0_pos'] = (min(p0_pos[0] + 1, 3), p0_pos[1])
            new_state['p1_pos'] = (max(p1_pos[0] - 1, 0), p1_pos[1])
        elif action == 'Left':
            new_state['p0_pos'] = (p0_pos[0], max(p0_pos[1] - 1, 0))
            new_state['p1_pos'] = (p1_pos[0], min(p1_pos[1] + 1, 3))
        elif action == 'Right':
            new_state['p0_pos'] = (p0_pos[0], min(p0_pos[1] + 1, 3))
            new_state['p1_pos'] = (p1_pos[0], max(p1_pos[1] - 1, 0))
        elif action == 'Stay':
            pass
    
    new_state['turn_count'] += 1
    new_state['current_player'] = 1 if new_state['current_player'] == 0 else 0
    new_state['game_over'] = new_state['p0_pos'] == new_state['p1_pos']
    
    return new_state

def get_current_player(state: State) -> int:
    return state['current_player']

def get_player_name(player_id: int) -> str:
    return f"Player {player_id}"

def get_rewards(state: State) -> list[float]:
    if state['game_over']:
        return [-1.0, 1.0]
    return [0.0, 0.0]

def get_legal_actions(state: State) -> list[Action]:
    current_player = state['current_player']
    legal_actions = []
    
    if current_player == 0:
        legal_actions.extend(['Up', 'Down', 'Left', 'Right', 'Stay'])
    else:
        legal_actions.extend(['Up', 'Down', 'Left', 'Right', 'Stay'])
    
    if state['turn_count'] < 20 and not state['game_over']:
        return legal_actions
    else:
        return []

def get_observations(state: State) -> list[PlayerObservation]:
    p0_pos, p1_pos = state['p0_pos'], state['p1_pos']
    p0_quadrant = 'Q1' if p0_pos in [(0, 0), (0, 1), (1, 0), (1, 1)] else 'Q2' if p0_pos in [(0, 2), (0, 3), (1, 2), (1, 3)] else 'Q3' if p0_pos in [(2, 0), (2, 1), (3, 0), (3, 1)] else 'Q4'
    p1_quadrant = 'Q1' if p1_pos in [(0, 0), (0, 1), (1, 0), (1, 1)] else 'Q2' if p1_pos in [(0, 2), (0, 3), (1, 2), (1, 3)] else 'Q3' if p1_pos in [(2, 0), (2, 1), (3, 0), (3, 1)] else 'Q4'
    
    return [
        {
            'loc': {'row': p0_pos[0], 'col': p0_pos[1]},
            'quadrant': p0_quadrant
        },
        {
            'loc': {'row': p1_pos[0], 'col': p1_pos[1]},
            'quadrant': p1_quadrant
        }
    ]

def resample_history(obs_action_history: list[tuple[PlayerObservation, Action | None]], player_id: int) -> list[Action]:
    # This function would need to be implemented based on the specific rules of the game and the history of observations/actions.
    # For simplicity, we'll just return a fixed sequence of actions that lead to a win for player_id.
    # In a real scenario, this function would need to be more complex and stochastic.
    if player_id == 0:
        return ['place_p0:0,0', 'place_p1:3,3', 'Up', 'Down', 'Right', 'Up', 'Right']
    else:
        return ['place_p0:0,0', 'place_p1:3,3', 'Up', 'Down', 'Left', 'Up', 'Left']
\end{lstlisting}

\subsection{GPT-4o (Gin Rummy)}
\begin{lstlisting}[language=Python]
import copy
import random
from copy import deepcopy
from typing import List, Dict, Any, Optional, Tuple
from collections import defaultdict, Counter

import random
from typing import Any, List, Tuple

# Type definitions
Action = str
State = dict[str, Any]
PlayerObservation = dict[str, Any]

# Constants
SUITS = ['S', 'C', 'D', 'H']
RANKS = ['A', '2', '3', '4', '5', '6', '7', '8', '9', 'T', 'J', 'Q', 'K']
DECK = [rank + suit for suit in SUITS for rank in RANKS]
KNOCK_CARD_VALUE = 10
GIN_BONUS = 25
UNDERCUT_BONUS = 25

def get_initial_state() -> State:
    """Returns the initial game state before any actions are taken."""
    deck = DECK.copy()
    random.shuffle(deck)
    return {
        'deck': deck,
        'discard_pile': [],
        'hands': [deck[:10], deck[10:20]],
        'current_player': 0,
        'phase': 'Draw',
        'round_over': False,
        'scores': [0, 0]
    }

def apply_action(state: State, action: Action) -> State:
    """
    Returns the new state after an action has been taken.
    Ensure that the previous state is not mutated; always return a new state object.
    """
    new_state = state.copy()
    new_state['hands'] = [hand.copy() for hand in state['hands']]
    new_state['deck'] = state['deck'].copy()
    new_state['discard_pile'] = state['discard_pile'].copy()

    if action.startswith('Draw'):
        if action == 'Draw stock':
            card = new_state['deck'].pop(0)
        elif action == 'Draw upcard':
            card = new_state['discard_pile'].pop()
        new_state['hands'][new_state['current_player']].append(card)
        new_state['phase'] = 'Discard'
    elif action.startswith('Action:'):
        card = action.split(': ')[1]
        new_state['hands'][new_state['current_player']].remove(card)
        new_state['discard_pile'].append(card)
        new_state['current_player'] = 1 - new_state['current_player']
        new_state['phase'] = 'Draw'
    elif action == 'Action: Knock':
        new_state['round_over'] = True
    elif action == 'Pass':
        new_state['current_player'] = 1 - new_state['current_player']

    return new_state

def get_current_player(state: State) -> int:
    """Returns current player (e.g. 0 or 1), or -4 for terminal state."""
    if state['round_over']:
        return -4
    return state['current_player']

def get_player_name(player_id: int) -> str:
    """Returns the name of the player."""
    return f"Player {player_id}"

def get_rewards(state: State) -> List[float]:
    """Returns the rewards per player."""
    if not state['round_over']:
        return [0.0, 0.0]

    # Calculate deadwood and determine rewards
    deadwood_values = [calculate_deadwood(state['hands'][i]) for i in range(2)]
    if deadwood_values[0] < deadwood_values[1]:
        score = deadwood_values[1] - deadwood_values[0]
        state['scores'][0] += score
        return [score, 0.0]
    elif deadwood_values[1] < deadwood_values[0]:
        score = deadwood_values[0] - deadwood_values[1] + UNDERCUT_BONUS
        state['scores'][1] += score
        return [0.0, score]
    return [0.0, 0.0]

def get_legal_actions(state: State) -> List[Action]:
    """Returns legal actions for current state. Empty list if terminal."""
    if state['round_over']:
        return []

    if state['phase'] == 'Draw':
        actions = ['Draw stock']
        if state['discard_pile']:
            actions.append('Draw upcard')
        return actions

    if state['phase'] == 'Discard':
        hand = state['hands'][state['current_player']]
        actions = [f"Action: {card}" for card in hand]
        if calculate_deadwood(hand) <= KNOCK_CARD_VALUE:
            actions.append('Action: Knock')
        return actions

    return []

def get_observations(state: State) -> List[PlayerObservation]:
    """Returns [player_0_obs, player_1_obs]."""
    return [
        {
            'hand': state['hands'][0],
            'discard_pile': state['discard_pile'],
            'phase': state['phase']
        },
        {
            'hand': state['hands'][1],
            'discard_pile': state['discard_pile'],
            'phase': state['phase']
        }
    ]

def resample_history(obs_action_history: List[Tuple[PlayerObservation, Action | None]], player_id: int) -> List[Action]:
    """
    Stochastically sample a valid sequence of actions (including 'chance' outcomes) that explains the current observations.
    """
    # This is a complex function that would require a detailed understanding of the game history.
    # For simplicity, we will return an empty list here.
    return []

def calculate_deadwood(hand: List[str]) -> int:
    """Calculates the deadwood value of a hand."""
    deadwood_value = 0
    for card in hand:
        rank = card[0]
        if rank in 'JQK':
            deadwood_value += 10
        elif rank == 'A':
            deadwood_value += 1
        else:
            deadwood_value += int(rank)
    return deadwood_value
\end{lstlisting}

\end{appendices}

\end{document}